\documentclass[11pt,a4paper]{article}
\usepackage[hyperref]{acl2020}
\usepackage{times}
\usepackage{latexsym}

\usepackage{helvet} % DO NOT CHANGE THIS
\usepackage{courier}  % DO NOT CHANGE THIS
\usepackage{graphicx}  % DO NOT CHANGE THIS
\usepackage{amsmath}
\usepackage{amsfonts}
\usepackage{multirow}
\usepackage{verbatim}
\usepackage[switch]{lineno}
\usepackage{microtype}
\aclfinalcopy
\def\aclpaperid{1146} %  Enter the acl Paper ID here

\title{Exploring Contextual Word-level Style Relevance for \\ Unsupervised Style Transfer}

\author{Chulun Zhou$^{1}$\footnotemark[2], \ Liangyu Chen$^{2}$\footnotemark[2], \ Jiachen Liu$^{2}$,\ Xinyan Xiao$^{2}$, \ \\ \textbf{Jinsong Su}$^{1}$\footnotemark[1]\textbf{,} \ \textbf{Sheng Guo}$^{2}$\textbf{,} \textbf{Hua Wu}$^{2}$\\
$^{1}$Xiamen University, Xiamen, China \ \ \ $^{2}$Baidu Inc., Beijing, China\\
 {\tt clzhou@stu.xmu.edu.cn jssu@xmu.edu.cn} \\ {\tt \{chenliangyu,liujiachen,xiaoxinyan,guosheng,wu\_hua\}@baidu.com}
}
\date{}

\begin{document}
\maketitle
\renewcommand{\thefootnote}{\fnsymbol{footnote}}
\footnotetext[2]{Equal contribution. This work is done when Chulun Zhou was interning at Baidu Inc., Beijing, China.}
\footnotetext[1]{Corresponding author}
\begin{abstract}
Unsupervised style transfer aims to change the style of an input sentence while preserving its original content without using parallel training data. In current dominant approaches, owing to the lack of fine-grained control on the influence from the target style, they are unable to yield desirable output sentences.
In this paper, we propose a novel attentional sequence-to-sequence (Seq2seq) model that dynamically exploits the relevance of each output word to the target style for unsupervised style transfer. 
Specifically, 
we first pre-train a style classifier, where the relevance of each input word to the original style can be quantified via layer-wise relevance propagation. 
In a denoising auto-encoding manner, we train an attentional Seq2seq model to reconstruct input sentences and repredict word-level previously-quantified style relevance simultaneously.
In this way, this model is endowed with the ability to automatically predict the style relevance of each output word.
Then, we equip the decoder of this model with a neural style component to exploit the predicted word-level style relevance for better style transfer.
Particularly, we fine-tune this model using a carefully-designed objective function involving style transfer, style relevance consistency, content preservation and fluency modeling loss terms.
Experimental results show that our proposed model achieves state-of-the-art performance in terms of both transfer accuracy and content preservation.
\end{abstract}

\section{Introduction}
Text style transfer is a task that changes the style of input sentences while preserving their style-independent content. 
Due to its wide applications, such as sentiment transfer \cite{DBLP:conf/icml/HuYLSX17,DBLP:conf/nips/ShenLBJ17} and text formalization \cite{DBLP:conf/aaai/JainMAS19}, 
it has become a research hotspot in natural language generation in recent years. 
However, 
due to the lack of parallel training data, 
researchers mainly focus on unsupervised style transfer.
    
%\begin{figure}[h]
%\includegraphics[scale=0.44]{Capture.JPG}
%\caption{
%A style transfer example produced by different models. 
%Baseline 1-4 are the recently proposed competitive models \cite{DBLP:conf/naacl/LiJHL18,DBLP:journals/corr/abs-1808-07894,DBLP:journals/corr/abs-1905-10060,DBLP:conf/acl/WuRLS19}, respectively. 
%We underline the parts of the input sentence that are highly relevant to its style and highlight the improperly generated parts of output sentences as red 
%while marking the reasonably transferred parts as blue. 
%}
%\label{fig:pic1}
%\end{figure} 

In this aspect, 
many approaches \cite{DBLP:conf/icml/HuYLSX17,DBLP:conf/nips/ShenLBJ17,DBLP:conf/aaai/FuTPZY18} resort to an auto-encoding framework, 
where the encoder is used to disentangle the content and style, 
and the decoder generates the output sentence with the target style. 
Another line of research \cite{DBLP:conf/acl/LiWZXRSZ18,DBLP:conf/naacl/LiJHL18} focuses on removing the style marker of the input sentence to obtain a style-independent sentence representation. 
When generating the output sentence, 
both lines directly feed the target style into the decoder as a whole. 
From another perspective, 
%Logeswaran et al. \shortcite{DBLP:conf/nips/LogeswaranLB18} apply auto-encoding with back-translation to solve the lack of parallel data. 
%Furthermore, 
some researchers treat the style transfer as a translation process and adapt unsupervised machine translation to this task \cite{DBLP:conf/nips/LogeswaranLB18,DBLP:journals/corr/abs-1808-07894,DBLP:conf/iclr/LampleSSDRB19},
where the style switch is implicitly achieved. 
Besides, 
many recent models further explore this task in different ways, 
such as gradient-based optimization method \cite{DBLP:journals/corr/abs-1905-12304}, dual reinforcement learning \cite{DBLP:journals/corr/abs-1905-10060}, hierarchical reinforcement learning \cite{DBLP:conf/acl/WuRLS19} and transformer-based model \cite{DBLP:conf/acl/DaiLQH19}. 
Overall, 
in these models, 
the quality of output sentences mainly depends on the content representation of input sentence and the exploitation of the target style.

%However, 
%one main drawback of the aforementioned models is that they lack the fine-grained control of the influence from the target style on the generation process, 
%resulting in unnecessary or incorrect modifications of many style-independent parts. 
%For example,
%in Figure \ref{fig:pic1},
%the output sentences of other models are either not fluent or deviate from their original content.
%In contrast,
%only our model, 
%which can dynamically model the influence of the target style on output words, 
%successfully changes the style of the input sentence while preserving original content well.
%Intuitively, 
%the frequencies of words occurring the sentences with different styles are distinct,
%and thus they are related to various styles in different degrees.
%In view of the above,
%we believe that during the ideal style transfer,
%impacts of target style information should be distinguished depending on different words. 
%However, existing models do not take this into account, 
%therefore limiting the potential of further style transfer improvement.

However, 
one main drawback of the aforementioned models is that they lack the fine-grained control of the influence from the target style on the generation process,
limiting the potential of further style transfer improvement.
Intuitively, 
the frequencies of words occurring the sentences with different styles are distinct,
and thus they are related to various styles in different degrees.
In view of the above,
we believe that during the ideal style transfer,
impacts of the target style should be distinguished depending on different words. 
If we equip the current style transfer model with a neural network component, 
which can automatically quantify the style relevance of the output sentence at word level, the performance of the model is expected to be further improved.

In this paper, 
we propose a novel attentional sequence-to-sequence model (Seq2seq) that dynamically predicts and exploits the relevance of each output word to the target style for unsupervised style transfer.
%Our basic intuition stems from the fact that 
%in one sentence, 
%words are related to the sentence style in different degrees.
%hen people change the style of an input sentence,
%they only need to modify those highly style-related words,
%while maintaining other style-independent words unchanged. 
%Thus, 
%we argue that if we equip the current style transfer model with a neural network component, 
%which can automatically quantify the style relevance of the output sentence at word level, 
%the performance of the model is expected to be further improved.
%To this end, 
%we first use a pre-trained classifier to quantify the relevance of each input word to the original style 
Specifically,
we first pre-train a style classifier,
where the relevance of each input word to the original style can be quantified through layer-wise relevance propagation (LRP) \cite{bach2015pixel}.
After that, 
in a denoising auto-encoding manner, 
we train a basic attentional Seq2seq model to reconstruct the input sentence and repredict its word-level previously-quantified style relevance simultaneously. 
In this way,
this model is endowed with the ability to automatically predict the style relevance of each output word.
Then,
we equip the decoder of this model with a neural style component to exploit the predicted word-level style relevance for better style transfer.
Particularly,
we fine-tune this model using a carefully-designed ojbective function involving style transfer, style relevance consistency, content preservation and fluency modeling loss terms.

Compared with previous approaches, 
our proposed model avoids the complex disentanglement procedure, 
of which the quality can not be guaranteed. 
Also, 
our model is able to solve the issue of the source-side information loss caused by unsatisfactory disentanglement or explicitly removing style markers. 
More importantly, our model is capable of achieving fine-grained control over the impacts of target style on different output words, 
leading to better style transfer.
To sum up, 
%Overall, our contributions in this work are threefold:
our contributions can be summarized as follows:
\begin{itemize}
\setlength{\itemsep}{3pt}
\setlength{\parsep}{0pt}
\setlength{\parskip}{0pt}
\item 
We explore a training approach based on LRP and denoising auto-encoding for the Seq2seq style transfer model, 
which enables the model to automatically predict the word-level style relevance of output sentences;
\item 
We propose a novel Seq2seq model, 
which exploits the predicted word-level style relevance of output sentences for better style transfer.
To the best of our knowledge, 
the text style transfer with fine-grained style control has not been explored before;
\item 
Experimental results and in-depth analysis on two benchmark datasets strongly demonstrate the effectiveness of our model. We release our code at \url{https://github.com/PaddlePaddle/Research/tree/master/NLP/ACL2020-WST}
\end{itemize}

\section{Related Work}
In recent years,
unsupervised text style transfer has attracted increasing attention. 
Most of previous work \cite{DBLP:conf/icml/HuYLSX17,DBLP:conf/nips/ShenLBJ17,DBLP:conf/aaai/FuTPZY18,DBLP:conf/acl/TsvetkovBSP18,DBLP:conf/acl/LiWZXRSZ18,DBLP:conf/naacl/LiJHL18} aimed at producing a style-independent content representation from the input sentence and generate the output with target style.
For example, 
Hu et al. \shortcite{DBLP:conf/icml/HuYLSX17} employed a variational auto-encoder with an attribute classifier as discriminator, 
forcing the disentanglement of specific attributes and content in latent representation. 
Shen et al. \shortcite{DBLP:conf/nips/ShenLBJ17} exploited an auto-encoder framework with an adversarial style discriminator to obtain a shared latent space cross-aligning the content of text from different styles. 
Based on multi-task learning and adversarial training of deep neural networks, 
Fu et al. \shortcite{DBLP:conf/aaai/FuTPZY18} explored two models to learn style transfer from non-parallel data.  Prabhumoye et al. \shortcite{DBLP:conf/acl/TsvetkovBSP18} learned a latent representation of the input sentence to better preserve the meaning of the sentence while reducing stylistic properties. Then adversarial training and multi-task learning techniques were exploited to make the output match the desired style. Although these work has shown effectiveness to some extent, however, as analyzed by some recent work \cite{DBLP:conf/emnlp/LiMSJRJ17,DBLP:conf/iclr/LampleSSDRB19}, 
their style discriminators are prone to be fooled. 

Meanwhile, 
some studies \cite{DBLP:journals/corr/abs-1905-10060,DBLP:conf/naacl/LiJHL18} explicitly removed style-related words identified by a pre-trained classifier to get a style-independent content representation, 
and then added target style to generate output sentences. 
Nevertheless, 
such approaches tend to cause the information loss of the input sentence, 
since its style-related words often contain meaningful content.
Besides, 
there have been several researches \cite{DBLP:conf/nips/LogeswaranLB18,DBLP:journals/corr/abs-1808-07894,DBLP:conf/iclr/LampleSSDRB19} adopting back-translation to build style transfer models without parallel data. 
Logeswaran et al. \shortcite{DBLP:conf/nips/LogeswaranLB18} introduced a reconstruction loss interpolating auto-encoding and back-translation loss components, 
where attribute compatibility is encouraged by a discriminator. 
Along this line, 
Zhang et al. \shortcite{DBLP:journals/corr/abs-1808-07894} and Lample et al. \shortcite{DBLP:conf/iclr/LampleSSDRB19} directly adapted unsupervised machine translation approaches to this task, 
where the style transfer is implicitly achieved via iterative back-translations between texts in different styles.

Very recently, 
some attempts \cite{DBLP:journals/corr/abs-1905-12304,DBLP:journals/corr/abs-1905-10060,DBLP:conf/acl/WuRLS19,DBLP:conf/acl/DaiLQH19} have been made to perform style transfer from other different perspectives. 
For example, Liu et al. \shortcite{DBLP:journals/corr/abs-1905-12304} mapped a discrete sentence into a continuous space and then used the gradient-based optimization with a pre-trained attribute predictor to find the latent representation satisfying desired properties (e.g. sentence length, sentiment). 
Luo et al. \shortcite{DBLP:journals/corr/abs-1905-10060} performed a one-step mapping to directly transfer the style of the original sentences via dual reinforcement learning. 
Wu et al. \shortcite{DBLP:conf/acl/WuRLS19} adopted a hierarchical reinforced sequence operation method to iteratively revise the words of original sentences. 
Dai et al. \shortcite{DBLP:conf/acl/DaiLQH19} proposed a Transformer-based \cite{DBLP:conf/nips/VaswaniSPUJGKP17} style transfer model without disentangling latent representation. Finally, note that exploring word-level style relevance has also been studied in other NLP tasks, such as machine translation \cite{zeng-etal-2018-multi, su2019exploring}.

\section{Our Model}
Given a set of labelled training instances $D=\{(X_i,s_i)\}_{i=1}^{|D|}$,
where $s_i$$\in$$\{0,1\}$ is the style label of the sentence $X_i$, 
we aim to train a style transfer model that can automatically convert an input sentence $X$=$x_1,x_2,...,x_{|X|}$ with the original style $s$ into a content-invariant output sentence $Y$=$y_1,y_2,...,y_{|Y|}$ with the target style $s'$.

\begin{figure*}[!htb]
\centering
\includegraphics[width=1\linewidth]{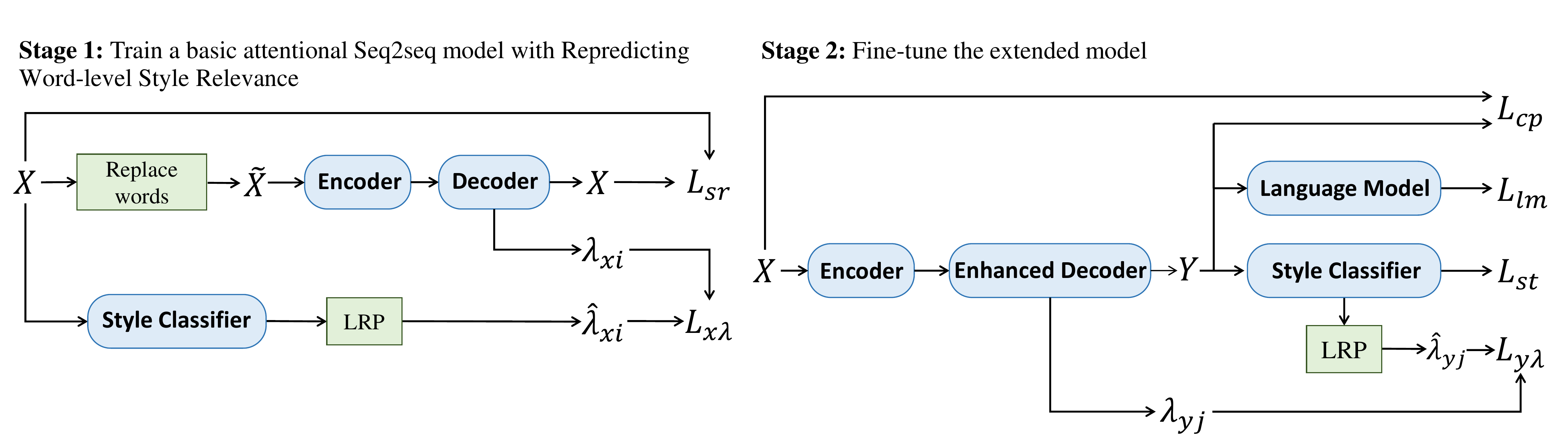}
\caption{
Two training stages of our model:
\textbf{Stage 1}: Train a basic attentional Seq2seq model that is required to reconstruct the input sentence and repredict its word-level style relevance; 
\textbf{Stage 2}: Extend the basic model by equipping its decoder with a neural style component, and then fine-tune it using a novel objective function. 
$X$ and $Y$ are the input sentence and the output sentence, respectively,
and $\widetilde{X}$ is the partially corrupted version of $X$.
LRP indicates layer-wise relevance propagation \cite{bach2015pixel}. 
$\lambda_{*}$ indicates the word-level style relevance.
$L_{*}$ represents the loss terms used in objective functions.
%The decoder and content reconstruction loss $L_{con}$ are two key components, 
%both of which are related to style-relevance $\lambda$ and used to precisely control transfer at word level. 
%Please note that the style classifier and language model in the figure are trained in advance.
}
\label{fig:overview}
\end{figure*}

To achieve this goal,
we extend the standard attentional Seq2seq model \cite{DBLP:conf/nips/SutskeverVL14,DBLP:journals/corr/BahdanauCB14} by equipping its decoder with a neural style component to achieve fine-grained control over the impacts of target style on different output words.
As shown in Figure \ref{fig:overview}, 
the training of our model consists of two stages,
%:
%(1) auto-encoding with style relevance restoration and
%(2) style transfer training.
%We describe them below.
which will be described below.

%\subsubsection{Stage 1: Auto-encoding with Style Relevance Restoration}
\subsection{Stage 1: Train a Basic Attentional Seq2seq Model with Repredicting Word-level Style Relevance}
At this stage, 
we first introduce a pre-trained style classifier to quantify the word-level relevance of training sentences to the original style via LRP \cite{bach2015pixel}.
Then, 
we train a basic attentional Seq2seq model in a denoising auto-encoding manner,
where this model is required to reconstruct the input sentence and repredict its word-level style relevance simultaneously.
By doing so, our model acquires the preliminary ability to predict the style relevance of output words and reconstruct input sentences, which makes the training in the subsequent stage easier.
Next, we briefly describe our basic model,
and then introduce its objective function in detail.

\subsubsection{Attentional Seq2seq Model}
It mainly consists of an encoder and a decoder with an attention mechanism.

The encoder is a forward GRU network.
Taking the sentence $X$=$\{x_1,x_2,...,x_{|X|}\}$ as input,
this network maps the input words into a hidden state sequence as %follows:
%\begin{equation}
$\mathbf{h}^{e}_{i}=\text{GRU}(\mathbf{e}(x_i),\mathbf{h}^{e}_{i-1})$,
%\end{equation}
where $\mathbf{e}(x_i)$ and $\mathbf{h}^{e}_{i}$ denotes the embedding vector and the hidden state of the word $x_i$, respectively. 
Specially, 
the last hidden state $\mathbf{h}^{e}_{|X|}$ is used to initialize the decoder hidden state $\mathbf{h}^{d}_0$.

The decoder is also a forward GRU network.
Its hidden state is updated by
%\begin{equation}\label{con:decoder_hidden1}
$\mathbf{h}^{d}_{j}=\text{GRU}(\mathbf{e}(y_{j-1}),\mathbf{h}^{d}_{j-1}, {\mathbf{c}_j})$,
%\end{equation}
where $\mathbf{h}^{d}_{j}$ is the decoder hidden state at the $j$-th timestep, $\mathbf{e}(y_{j-1})$ is the embedding vector of the previously generated word $y_{j-1}$, and $\mathbf{c}_j$ denotes the corresponding attention-based context vector. 
Formally, 
$\mathbf{c}_j$ is defined as the weighted sum of all hidden states of input words:
\begin{align}\label{Eqa_AttenContVector}
\mathbf{c}_j&=\sum_{i}^{|X|}\frac{\text{exp}(e_{ji})}{\sum_{i'=1}^{|X|}{\text{exp}(e_{ji'})}}\mathbf{h}^e_{i}, \\
e_{ji}&=\mathbf{v}^{\top}\text{tanh}(\mathbf{W_e}\mathbf{h}^{e}_{i} + \mathbf{W_d}\mathbf{h}^{d}_{j}),
\end{align}
where $\mathbf{v}$, $\mathbf{W_e}$, $\mathbf{W_d}$ are trainable parameters.
Finally,
the output prediction probability over vocabulary is calculated as $P(y_j|y_{<j},X)=Softmax(\mathbf{W}_{o}\mathbf{h}^{d}_{j})$,
%\begin{equation}\label{con:output}
%P(y_j|y_{<j},X)=\text{softmax}(\mathbf{W}_{o}\mathbf{h}^{d}_{j}),
%\end{equation}
where $\mathbf{W}_o$ is a learnable matrix.
Please note that all bias terms in above equations are also trainable parameters,
which are omitted for the simplicity of notation.

\subsubsection{The Objective Function}
To effectively train the above basic model,
we define the following objective function $L_1$:
\begin{align}\label{Eqa_StageLossFun1}
L_1 = L_{sr}+L_{x\lambda},
\end{align}
where $L_{sr}$ and $L_{x\lambda}$ denote 
the sentence reconstruction loss and 
the style relevance restoration loss,
respectively.

1. \textbf{Sentence reconstruction loss} $L_{sr}$:
Using this loss, 
we expect our model to capture informative features for reconstructing sentence.
Formally,
we define $L_{sr}$ as follows:
\begin{equation}
L_{sr}(\theta_{s2s})=-\sum_{i=1}^{|\widetilde{X}|}{\text{log}P(x_{i}|x_{<i}, \widetilde{X})}.
\end{equation}
Here, 
$\theta_{s2s}$ denotes parameters of this Seq2seq model,
and $\widetilde{X}$ is the partially corrupted version of $X$, 
%where input words are randomly replaced according to a certain rate, 
%aiming to prevent our model from simply copying $X$.
where a certain proportion of input words are randomly replaced to prevent our model from simply copying $X$.

2. \textbf{Style relevance restoration loss} $L_{x\lambda}$:
It is used to measure how well the word-level style relevance of an input sentence can be repredicted during the denoising auto-encoding.
Formally,  
it is defined as
\begin{equation}\label{Eqa_SRRE}
L_{x\lambda}(\theta_{s2s}, \theta_{\lambda})
=\frac{1}{|X|}\sum_{i=1}^{|X|}{(\lambda_{xi}-\hat{\lambda}_{xi})^2},
\end{equation}
where $\lambda_{xi}\in(0,1)$ and $\hat{\lambda}_{xi}\in(0,1)$ denote the style relevance of the $i$-th input and output word, respectively,
and $\theta_{\lambda}$ denotes the set of other parameters used to calculate $\hat{\lambda}_{xi}$ (see Equation \ref{Eqa_PredStyleRele}). It is notable that ${\hat{\lambda}_{xi}}$ and ${\hat{\lambda}_{xi}}$ are not involved into the sentence reconstruction.
%, which will be explained later.
%Here, 
%the values of both $\lambda_{xi}$ and $\hat{\lambda}_{xi}$ are between 0 and 1.
Apparently, two key issues arise, 
namely, 
how to calculate $\lambda_{xi}$ and $\hat{\lambda}_{xi}$?

As for $\hat{\lambda}_{xi}$,
we calculate it based on the previous decoder hidden state $\textbf{h}^d_{i-1}$:
\begin{equation}\label{Eqa_PredStyleRele}
\hat{\lambda}_{xi}=\mathbf{v}_{\lambda}^{\top}\text{tanh}(\mathbf{W}_{\lambda}\mathbf{h}^{d}_{i-1}),
\end{equation}
where 
$\mathbf{W}_{\lambda}$, $\mathbf{v}_{\lambda}$ form the previously-mentioned parameter set $\theta_{\lambda}$ (see Equation \ref{Eqa_SRRE}).

To obtain $\lambda_{xi}$,
we use the instances in the training set $D$ to pre-train a TextCNN \cite{DBLP:conf/emnlp/Kim14} style classifier.
During this process,
we apply LRP \cite{bach2015pixel},
which has been widely used to measure the contributions of neurons to the final prediction of a classifier,
to quantify the word-level style relevance of sentences.
Concretely,
we calculate the relevance score $r^{(l)}_{k}$ of the $k$-th neuron $n^{(l)}_{k}$ at the $l$-th layer in a manner similar to back-propagation:
\begin{align}\label{eq:12}
r_{k}^{(l)} &= {\sum_{k'=1}^K{\frac{z_{kk'}}{\sum_{k''}{z_{kk''}}}}r_{k'}^{(l+1)}},
\end{align}
where $z_{kk'} = v_{k}^{(l)}w_{kk'}^{(l,l+1)}$.
Here, $w_{kk'}^{(l,l+1)}$ denotes the weight of the edge between adjacent neurons, 
$v^{(l)}_k$ is the value of neuron $n^{(l)}_k$ that can be computed in advance during the forward-propagation, and $K$ denotes the neuron number of each layer.
Through this way, 
we can obtain the neuron-wise contribution scores $\{r^{(0)}_{k}(x_i)\}^{K}_{k=1}$ of the embedding vector of input word $x_i$ to the final style prediction. 
Furthermore, 
we define the style relevance score $r(x_i)$ of $x_i$ as the sum of  $\{r^{(0)}_{k}(x_i)\}^{K}_{k=1}$,
and finally map this score into the range $[0,1)$ via a \emph{tanh}(*) function:
%, obtaining the style relevance
$\lambda_{xi}$$=$$\text{tanh}(\eta{|r(x_i)}|)$,
where the hyper-parameter $\eta$ serves as a scaling factor. 
In practice, 
since too low style relevance may be noise, 
we directly treat the words with style relevance lower than $\epsilon$ as style independent and set their style relevance as 0.\footnote{Experiments show that such treatment enhances the stability of our system.}

\begin{figure}[t]
\centering
\includegraphics[scale=0.44]{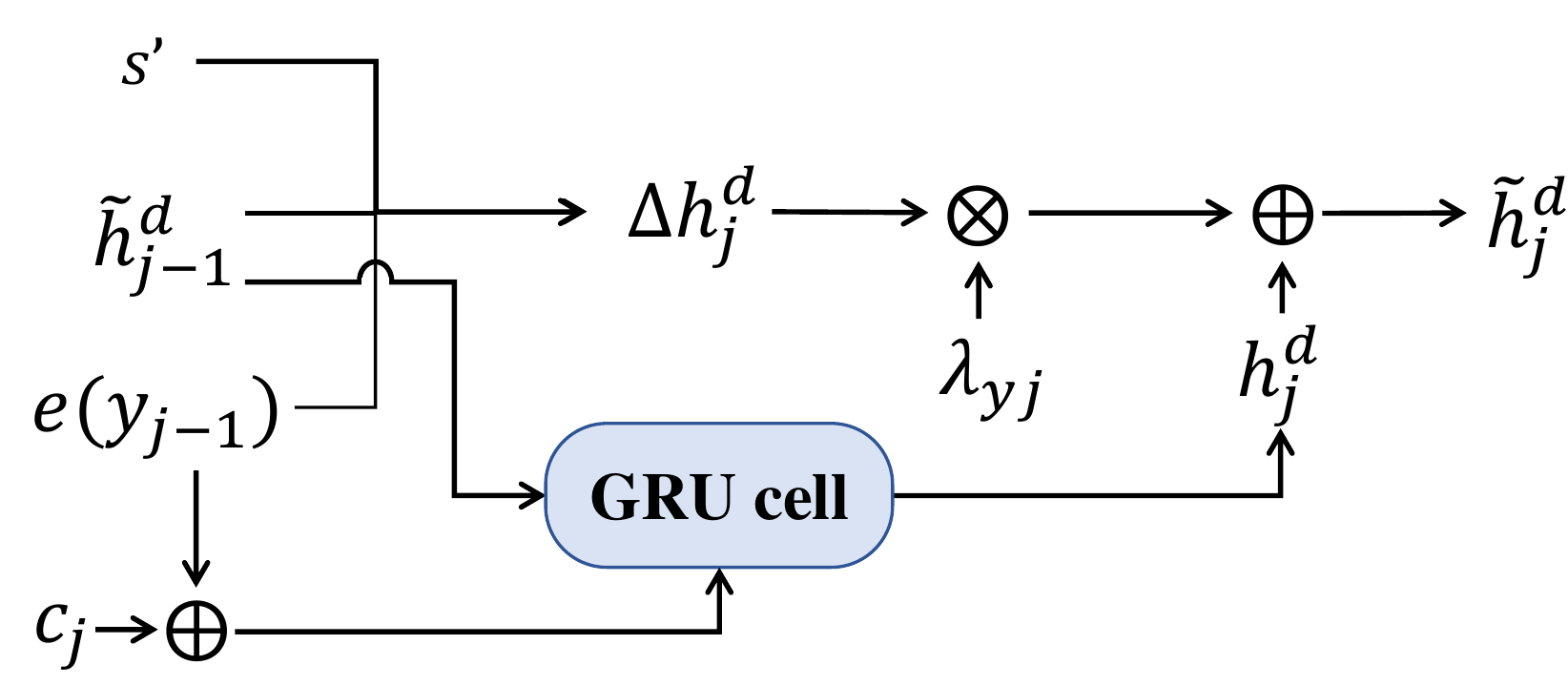}
\caption{
The architecture of our proposed neural style component. $\Delta\mathbf{h}^{d}_j$ is the revision to the hidden state $\mathbf{h}^{d}_{j}$, $\lambda_{yj}$ is a gate controlling to what extent the $\mathbf{h}^{d}_{j}$ will be revised by $\Delta\mathbf{h}^{d}_j$ and $\Tilde{\mathbf{h}}^{d}_{j}$ is the revised hidden state used to generate ${y}$.
}
\label{fig:stylec}
\end{figure}

%\subsection{Stage 2: Style Transfer Training}
\subsection{Stage 2: Fine-tune the Extended Model}
At this stage,
we extend the above basic model by equipping its decoder with a neural style component,
which predicts and then exploits the style relevance of the next output word to refine its generation.
Likewise, 
we first give a detailed description to our extended model,
and then depict how to fine-tune it using a novel objective function involving multiple loss terms.

\subsubsection{The Extended Model}
Here, we omit the descriptions of our basic encoder and decoder, 
which are identical to those of the previously-described basic model,
and only depict the newly introduced neural style component.
Figure \ref{fig:stylec} shows the architecture of this component.

%The dotted box of Figure \ref{fig:overview} provides the architecture of our component.
To incorporate word-level style relevance into our decoder,
at the $j$-th timestep, 
we predict the style relevance $\lambda_{yj}$ following Equation \ref{Eqa_PredStyleRele},
and then use $\lambda_{yj}$ to revise the decoder hidden state $\mathbf{h}^{d}_{j}$ as follows:
\begin{equation}
\label{Eqa_HSRevFunc}
\Tilde{\mathbf{h}}^{d}_{j}=\mathbf{h}^{d}_{j}+\lambda_{yj}\Delta\mathbf{h}^{d}_{j},
\end{equation}
where 
$\Delta\mathbf{h}^{d}_j$ is the revision to the hidden state $\mathbf{h}^{d}_{j}$, $\lambda_{yj}$ is a gate controlling to what extent the $\mathbf{h}^{d}_{j}$ will be revised by $\Delta\mathbf{h}^{d}_j$ and $\Tilde{\mathbf{h}}^{d}_{j}$ is the revised hidden state used to generate ${y_j}$ (see Subsection 2.1.1). Note that $\lambda_{yj}$ reflects how much information of target style is incorporated. 

More specifically,
$\mathbf{h}^{d}_{j}$ and $\Delta\mathbf{h}^{d}_{j}$ are updated as below:
\begin{align}
\mathbf{h}^{d}_{j}&=\text{GRU}(\mathbf{e}(y_{j-1}),\Tilde{\mathbf{h}}^{d}_{j-1}, \mathbf{c}_j), \\
\Delta\mathbf{h}^{d}_{j}&=f(\mathbf{e}(y_{j-1}),\Tilde{\mathbf{h}}^{d}_{j-1},s';\theta_{\Delta h}),
\end{align}
where 
$\mathbf{e}(y_{j-1})$ is the embedding vector of $y_{j-1}$,
$\mathbf{c}_j$ is the context vector calculated as Equation \ref{Eqa_AttenContVector},
$s'$ is the target style,
and 
$f(*)$ is an MLP function with the parameter set $\theta_{\Delta h}$ used to produce $\Delta\mathbf{h}^{d}_{j}$.

%Please note that the above revision to the decoder hidden %state is very important 
%because it 
By doing so, 
we can impose fine-grained control over the influence of the target style on the generation of the next word.
When our model predicts that the next word is strongly style-relevant,
it encourages the decoder to produce the proper stylistic content by leveraging more target style information.
Conversely,
%if the next word is predicted to be little relevant to the style, 
the target style will exert less influence on the decoder hidden state,
%getting rid of the improper disturbance to the generation of the word.
avoiding being disturbed to lose original style independent content.

\subsubsection{The Objective Function}
After initializing our extended model with parameters of the basic model,
we then fine-tune it using the following objective function:
\begin{equation}
\label{Eqa_StageLossFun2}
L_{2} = L_{st} + \alpha L_{y\lambda} + \beta L_{cp} + \gamma L_{lm},
\end{equation}
where $L_{st}$, $L_{y\lambda}$, $L_{cp}$ and $L_{lm}$ denote the style transfer loss, the style relevance consistency loss, the content preservation loss, and the frequency modeling loss,
with $\alpha, \beta, \gamma$ as their balancing parameters.

%\subsubsection{Style Conversion}
1. \textbf{Style transfer loss} $L_{st}$:
It is used to ensure the output sentence contains the target style well.
To this end, 
we apply the above-mentioned pre-trained style classifier (See \textbf{Stage 1}) to classify the style of the output sentence,
where related parameters of our model are updated to encourage the target style can be predicted from the output sentence:
\begin{equation}\label{con:L_trans}
L_{st}(\theta_{s2s},\theta_{\lambda},\theta_{\Delta h})
=-\mathbb{E}_{({X},s)\sim{D}} [\text{log}P(s'|G(Y)].
\end{equation}
$G(Y)$ denotes a ``soft'' generated sentence based on gumbel-softmax distribution \cite{DBLP:conf/iclr/JangGP17},
where the representation of each word is defined as the weighted sum of word embeddings with the prediction probability at the current timestep.

2. \textbf{Style relevance consistency loss} $L_{y\lambda}$:
It ensures the predicted style relevances of output words are consistent with its stylistic outcomes evaluated by the classifier. Specifically, during the above style classification,
we apply LRP to obtain the style relevance $\hat{\lambda}_{yj}$ of each ``soft'' word,
and then try to minimize the following loss:
\begin{equation}
L_{y\lambda}(\theta_{s2s}, \theta_{\lambda})=\frac{1}{|Y|}\sum_{j=1}^{|Y|}{(\lambda_{yj}-\hat{\lambda}_{yj})^2}.
\end{equation}

3. \textbf{Content preservation loss} $L_{cp}$:
However, 
only using the above two style-related loss terms will lead to model collapse, 
producing an extremely short output sentence 
that matches the target style but totally loses the original meaning.
%(e.g., ``\emph{the salads were fresh and crispy.}'' $\rightarrow$ ``\emph{horrible service!}'').
To address this issue, 
we introduce a content preservation loss $L_{cp}$ to prevent the model from collapsing.
    
Specifically,
we define the content representations of input sentence $X$ and output sentence $Y$ as the weighted sums of their individual word embeddings according to the corresponding weights $1-|\lambda_{xi}|$ and $1-|\lambda_{yj}|$, respectively.
Then, 
we minimize the following loss term to force these two representations to be close:
\begin{align}\label{con:L_cont}
&L_{cp}(\theta_{s2s},\theta_{\lambda},\theta_{\Delta h}) \\
=&[\sum_{i}^{|X|}{(1-|\lambda_{xi}|)\mathbf{e}(x_i)}   -\sum_{j}^{|Y|}{(1-|\lambda_{yj}|)\mathbf{e}(y_j)}]^2. \notag
\end{align}
In this way,
the less relevant a word is to the corresponding style, the more its embedding should be considered.

%\subsubsection{Fluency Modeling}
4. \textbf{Fluency modeling loss} $L_{lm}$:
Finally,
we follow Yang et al. \shortcite{DBLP:conf/nips/YangHDXB18} to introduce a bidirectional GRU based language model, 
which is pre-trained on the training instances with target style to ensure that our model can generate fluent output sentences.

For the forward direction,
we aim to reduce the distribution divergence between the prediction probability vector of our model and that of the forward language model by minimizing their cross-entropy as follows:
\begin{equation}
\begin{aligned}
\label{con:forwardlmloss}
\overrightarrow{L}\!_{lm}(\theta\!_{s2s},\!\theta\!_{\lambda},\!\theta\!_{\Delta h})
\!=\!\sum_{j=1}^{|Y|}\textbf{P}(*|y_{\!<j},\!X)\!^{\top}\!\text{log}(\!\overrightarrow{\!\textbf{P}}\!(*|y_{\!<j})),\!
\end{aligned}
\end{equation}
where $\textbf{P}(*|y_{<j},X)$ and $\overrightarrow{\textbf{P}}(*|y_{<j})$ are the predicted probability distributions produced by our model and the forward language model, respectively. 
Note that at each timestep, 
we fed the continuous approximation of the output word, 
which is defined as the weighted sum of word ebmeddings with the current probability vector, 
into the language model.
For the back direction,
we directly reverse the output sentence and calculate the reverse language model loss $\overleftarrow{L}_{lm}$ in a similar way.
Finally, 
we directly define the total fluency modeling loss $L_{lm}$ as the average of $\overrightarrow{L}_{lm}$ and $\overleftarrow{L}_{lm}$.

\section{Experiment}
%We conduct experiments on two benchmark datasets, which are used for sentiment transfer and  formality transfer, respectively.

\begin{table*}
\small
\begin{center}
\begin{tabular}{l|llll|llll}
\hline
\multicolumn{1}{c|}{\multirow{2}{*}{Model}} & \multicolumn{4}{c|}{YELP}         & \multicolumn{4}{c}{GYAFC} \\\cline{2-9}
\multicolumn{1}{r|}{}                  & \multicolumn{1}{r}{Acc} & BLEU & G2 & H2 & Acc     & BLEU   & G2    & H2    \\ \hline
CrossAlign \cite{DBLP:conf/nips/ShenLBJ17}   &   75.3                                &        17.9               &  36.7 & 28.9  & 70.5  &  3.6     &     15.9 &  6.8        \\
DelRetri \cite{DBLP:conf/naacl/LiJHL18}                                      &                89.0       &  31.1  & 52.6  & 46.1  & 55.2      &  21.1    &  34.2    & 30.6     \\
Unpaired \cite{DBLP:conf/acl/LiWZXRSZ18}                                      &             64.9          & 37.0  & 49.0  & 47.1  &  79.5     &  2.0    & 12.6     & 3.9 \\
UnsuperMT \cite{DBLP:journals/corr/abs-1808-07894}  &  95.4 &       44.5  &  65.1 & 60.7  & 70.8  &  33.4     &     48.6 &  45.4        \\
DualRL \cite{DBLP:journals/corr/abs-1905-10060}  &  85.6  &  55.2              &  68.7 & 67.1 & 71.1  & 41.9     &     54.6 &  52.7        \\
PoiGen \cite{DBLP:conf/acl/WuRLS19}  &  91.5  &  59.0  &  73.5 & 71.8  & 46.2  &  45.8     &    46.0 &  46.0
\\ \hline
Our Model    &      94.0                 &  \textbf{60.4}   &  \textbf{75.4}&  \textbf{73.6} &    \textbf{81.4}   &      \textbf{47.7} &   \textbf{62.3}   &  \textbf{65.2}   \\ \hline
\end{tabular}
\caption {Performance of different models in YELP and GYAFC. 
\emph{Acc} measures the percentage of output sentences that match the target style. 
\emph{BLEU} measures the content similarity between the output and the corresponding four human references. 
\emph{G2} and \emph{H2} denotes the geometric mean and harmonic mean of \emph{Acc} and \emph{BLEU}, respectively. Numbers in bold mean that the improvement
to the best performing baseline is statistically significant (t-test with p-value \textless 0.05).
}
\label{tab:performance_main}
\end{center}
\end{table*}

\begin{table*}
\small
\begin{center}
\begin{tabular}{l|llll|llll}
\hline
\multicolumn{1}{c|}{\multirow{2}{*}{Model}} & \multicolumn{4}{c|}{YELP}         & \multicolumn{4}{c}{GYAFC} \\  
\cline{2-9}
\multicolumn{1}{r|}{}                  & \multicolumn{1}{r}{Acc} & Con & Flu & Avg  & Sty    & Con   & Flu   & Avg   \\ 
\hline
DelRetri \cite{DBLP:conf/naacl/LiJHL18}                                      &               2.18       &  2.21  & 2.40  & 2.26    &  1.53   &  1.55   & 1.62  &  1.57   \\
UnsuperMT \cite{DBLP:journals/corr/abs-1808-07894}   &    3.26                                &       3.07           &  3.24 & 3.19  &  2.46    &     2.42 &  2.75     &  2.54\\
DualRL \cite{DBLP:journals/corr/abs-1905-10060}   &  3.31                           &        3.43           &  3.47 & 3.40  &2.26     &     2.28  &  2.36     & 2.30 \\
PoiGen \cite{DBLP:conf/acl/WuRLS19}   &  3.42                           &      3.51              &  3.54 & 3.49  & 1.39   &    1.52 & 1.43  &   1.45
\\ \hline
Our Model &     3.38               & \textbf{3.70} &  3.66 & \textbf{3.58}  & \textbf{2.91}    &    \textbf{3.00}  &  \textbf{3.14}  &   \textbf{3.02} \\ 
\hline
\end{tabular}
\caption {Human evaluation results. 
We show human rating (1-5) for transfer accuracy (\emph{Acc}), content preservation (\emph{Con}), fluency (\emph{Flu}). The average ratings (\emph{Avg}) are also calculated as overall scores. Numbers in bold mean that the improvement
to the best performing baseline is statistically significant (t-test with p-value \textless 0.05).
}
\label{tab:performance_human}
\end{center}
\end{table*}

\subsection{Dataset}
\textbf{YELP:}
This dataset is comprised of restaurant and business reviews and has been widely used in sentiment transfer.
To evaluate our model, 
we adopt the human references released by 
Luo et al. \shortcite{DBLP:journals/corr/abs-1905-10060}, 
which has four references for each sentence in the test set. 
Following common practices \cite{DBLP:conf/nips/ShenLBJ17,DBLP:conf/naacl/LiJHL18}, 
we choose reviews over 3 stars as positive instances and those under 3 stars as negative instances. 
The splitting of train, development and test sets are in accordance with the setting in \cite{DBLP:conf/naacl/LiJHL18}. 
Moreover, 
we filter the sentences with more than 16 words,
leaving roughly 448K, 64K and 1K sentences in the train and development and test sets, respectively.

\noindent\textbf{GYAFC:} The parallel data of GYAFC \cite{DBLP:conf/naacl/RaoT18} consists of formal and informal texts while providing four human references for each test sentence. 
Particularly, 
we use this dataset in a non-alignment setting during training. There are roughly 102K, 5K and 1K sentences remaining in the train, development and test sets, respectively.

\subsection{Baselines}
We compare our model with several competitive baselines: \textbf{CrossAlign} \cite{DBLP:conf/nips/ShenLBJ17}, \textbf{DelRetri} \cite{DBLP:conf/naacl/LiJHL18}, \textbf{Unpaired} \cite{DBLP:conf/acl/LiWZXRSZ18}, \textbf{UnsuperMT} \cite{DBLP:journals/corr/abs-1808-07894}, \textbf{DualRL} \cite{DBLP:journals/corr/abs-1905-10060}, \textbf{PoiGen} \cite{DBLP:conf/acl/WuRLS19}. 
Among these models, 
\emph{DualRL} and \emph{PoiGen} 
%are the recently proposed models exhibiting 
exhibit the best performance. 
%Specially, 
%we cite all the baseline results from \cite{DBLP:journals/corr/abs-1905-10060}.

 \subsection{Training Details}
As for the threshold $\epsilon$ for filtering noise, we set it to 0.3 by observing model performances on development set. For the training of Stage 2,
we empirically set the learning rate to $1\times{10^{-5}}$ and 
%the gradients are clipped if its norm exceeds $10^{-2}$.
clip the gradients if their norms exceed $10^{-2}$.
We draw $\alpha$ from 0.5 to 1.5 with the step 0.1, $\beta$ from 0.5 to 5 with the step 0.5 and $\gamma$ from 0.1 to 2 with the step 0.1.
Similarly, the overall performance of our model on the developement set is employed to guide the hyperparameter search procedure. Finally, we choose $\alpha=1$, $\beta=2$ and $\gamma=0.5$.

\subsection{Automatic Evaluation}
We evaluate the quality of output sentences in terms of transfer accuracy and content preservation.
Following previous work %\cite{DBLP:conf/naacl/LiJHL18,DBLP:journals/corr/abs-1808-07894,DBLP:journals/corr/abs-1905-10060},
\cite{DBLP:journals/corr/abs-1905-10060}, 
we use the pre-trained style classifier to calculate the transfer accuracy of output sentences. 
The classifier achieves 97.8\% and 88.3\% accuracy on the test sets of YELP and GYAFC, respectively. 
Moreover, 
we compute the BLEU scores of output sentences to measure content preservation.
%by referring to the human references provided by \cite{DBLP:journals/corr/abs-1905-10060}. 
Finally, we 
%follow Luo et al. \shortcite{DBLP:journals/corr/abs-1905-10060} to 
report the geometric mean and harmonic mean of these metrics, which quantify the overall performance of various models.

Experimental results in Table \ref{tab:performance_main} shows that our model achieves the best performance among all models.

\subsection{Human Evaluation}
We invite 3 annotators with linguistic background to evaluate the output sentences.\footnote{We use Fleiss kappa to quantify the agreement among them. The Fleiss kappa score is 0.76 for the Yelp dataset and 0.80 for the GYAFC dataset.}
%generated by different models. 
%For fair evaluation, 
%they are not aware of the source model of each output sentence. 
The accuracy of style transfer (\emph{Acc}), 
the preservation of original content (\emph{Con}) and the fluency (\emph{Flu}) are the three aspects of model performance we consider.
Following the criteria introduced in \cite{DBLP:journals/corr/abs-1808-07894}, the annotators are required to score each aspect of sentences from 1 to 5.

Table \ref{tab:performance_human} shows the human evaluation results.
%of different models. 
Our model achieves the best performance on both datasets in terms of almost every aspects, 
except that the \emph{Acc} score of our model is slightly lower than \emph{PoiGen} on YELP. 
It may be due to the error of the pre-trained style classifier that the transfer accuracy of our model is higher than \emph{PoiGen} on YELP.
%in automatic evaluation. 
%It is noticeable 
Note that the content preservation of our model is significantly higher than others, 
showing our word-level control actually preserves more style-independent content of original sentences.

\begin{table}[]
\begin{center}
\small
\begin{tabular}{l|llll}
\hline
\multicolumn{1}{c|}{Model}   & \multicolumn{1}{r}{Acc} & BLEU & G2   & H2   \\ 
\hline
Our Model     & 94.0 & 60.4 & \textbf{75.4} & \textbf{73.6} \\ \hline
\ \ \ \ \ \textit{-NSC}         & 88.3 & 55.7 & 70.1 & 68.3 \\
\ \ \ \ \ \textit{NSC-$\lambda_{yj}$} & \textbf{98.1} & 10.1 & 31.5 & 18.3 \\
\ \ \ \ \ \textit{-$L_{x\lambda}$} & 95.2 & 57.7 & 74.1 & 71.8 \\
\ \ \ \ \ \textit{$L_{cp}$$\rightarrow$$L'_{cp}$} & 78.2 & \textbf{61.6} & 69.4 & 68.9 \\
\ \ \ \ \ \textit{-$L_{y\lambda}$} & 94.3 & 59.4 & 74.8 & 72.8 \\
\ \ \ \ \ \textit{-$L_{lm}$} & 96.8 & 56.5 & 73.9 & 71.3 \\
\ \ \ \ \ \textit{Finetuning-} & 91.5 & 55.8 & 71.4 & 68.9\\
%\ \ \ \ \ \textit{-NSC \&\& $L_{cp}$$\rightarrow$$L'_{cp}$}     & 73.1 & 61.1 & 66.8 & 66.6 \\
\hline
\end{tabular}
\caption{Ablation study results. 
%\textit{-NSC} is the variant of our model without the proposed neural style component. 
%$L_{cp}$$\rightarrow$$L'_{cp}$ is the variant of our model replacing $L_{cp}$ (See Equation \ref{con:L_cont}) with $L'_{cp}$ (See Equation \ref{con:L_cont2}) in the objective function.
}
\label{tab:Ablation}
\end{center}
\end{table}

\begin{figure*}
\centering
\includegraphics[width=1 \linewidth]{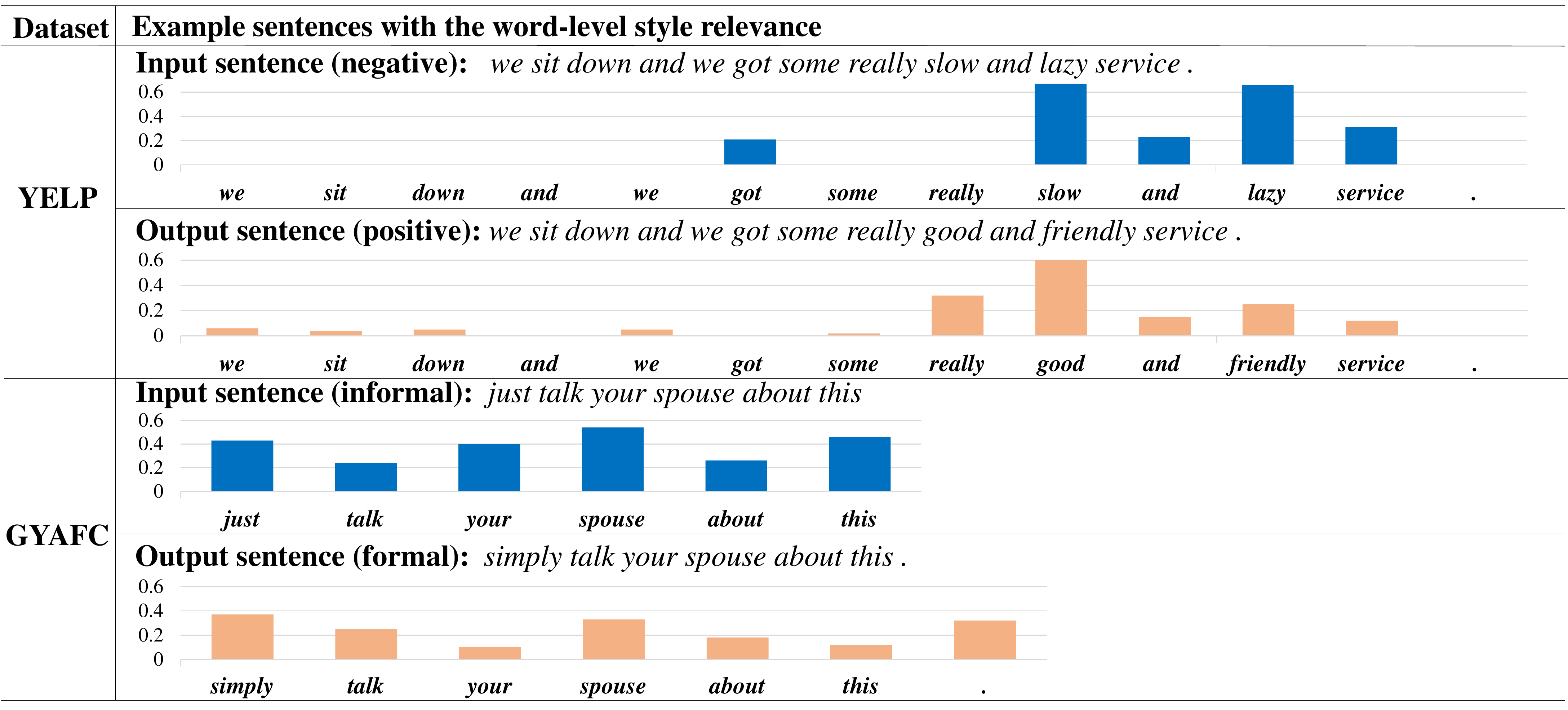}
\caption{
Sampled outputs of our model. Note that our model successfully transfers the style of these input sentences by not only replacing but also adding tokens. We also present the word-level style relevance of all sentences.
}
\label{fig:output}
\end{figure*}

\subsection{Ablation Studies}
%To investigate the impacts of $L_{cp}$ and our proposed neural style component (See \textbf{The Extended Model of Stage 2}),
%both of which play important roles in exploiting the word-level style relevance to the target style,
%we conduct ablation experiments on YELP that is larger than GYAFC. 
%Specifically,
%we compare our model with its three variants:
Compared with previous studies, 
the training of our model contains two stages, 
involving a neural style component and several novel loss terms, 
such as $L_{x\lambda}$ (see Equation \ref{Eqa_StageLossFun1}), $L_{y\lambda}$, $L_{cp}$ and $L_{lm}$ (see Equation \ref{Eqa_StageLossFun2}).
To fully investigate their effects on our model,
we conduct extensive ablation studies on YELP,
which is larger than GYAFC.
Specifically,
we compare ours with the following variants:
\begin{itemize}
\setlength{\itemsep}{3pt}
\setlength{\parsep}{0pt}
\setlength{\parskip}{0pt}
\item
-\textit{NSC}:
A variant of our model, 
where the proposed \textbf{n}eural \textbf{s}tyle \textbf{c}omponent is removed from the model.
It should be noted that this variant is actually the basic model,
which is only trained at Stage 1.
\item
\textit{NSC}-$\lambda_{yj}$:
It is a variant of our model,
which is equipped with a neural style component but without $\lambda_{yj}$ (see Equation \ref{Eqa_HSRevFunc}).
Note that this variant does have the ability of fine-grained controlling the influence from the target style on the generation process.
\item
-$L_{x\lambda}$:
%During the training Stage 1 of this variant, 
%the loss term $L_{x\lambda}$ is directly removed from Equation \ref{Eqa_StageLossFun1}.
In this model,
the loss term $L_{x\lambda}$ is directly removed from Equation \ref{Eqa_StageLossFun1} at Stage 1.
\item
$L_{cp}$$\rightarrow$$L'_{cp}$\footnote{Because our model would collapse if $L_{cp}$ is completely removed from the objective function, 
we do not compare our model with its variant without $L_{cp}$ loss.}:
A variant of our model, 
where its content preservation loss is modified as 
\begin{align}\label{con:L_cont2}
L'_{cp}(\theta_{s2s},\!\theta_{\lambda},\!\theta_{\Delta h})\!
=\!(\sum_{i}^{|X|}\mathbf{e}(x_i) \!-\!\sum_{j}^{|Y|}\mathbf{e}(y_j))^2.
\end{align}
Compared $L'_{cp}$ with $L_{cp}$ (see Equation \ref{con:L_cont}),
we find that all words are equally considered in $L'_{cp}$.
Thus, through this experiment,
we can investigate the impact of differential word-level style relevance modeling on our model. 
\item -$L_{y\lambda}$: 
It is also a variant of our model,
where the weight $\alpha$ of $L_{y\lambda}$ is set as 0 (see Equation \ref{Eqa_StageLossFun2}).
\item
-$L_{lm}$: 
A variant of our model with the weight $\gamma$ of $L_{lm}$ as 0 (see Equation \ref{Eqa_StageLossFun2}).
\item
\textit{Finetuning}-:
For this model, 
we fix all parameters involved in Stage 1 at Stage 2.
%\item
%-\textit{NSC} \&\& $L_{cp}$$\rightarrow$$L'_{cp}$:
%a variant of our model, 
%where we not only replace $L_{cp}$ with $L'_{cp}$, 
%but also remove the neural style component from our model.
\end{itemize}

%The results are shown in Table \ref{tab:Ablation}. 
%We have the following observations:
%\textbf{First}, 
%if we only remove the newly introduced neural style component,
%both the \emph{Acc} and \emph{BLEU} scores of our model drop a lot.
%This result strongly verifies the effectiveness of our neural style component.
%\textbf{Second},
%when we replace the loss term $L_{cp}$ with $L'_{cp}$,
%the \emph{Acc} score of our model significantly drops but the \emph{BLEU} score increases.
%The underlying reason may be with $L_{cp}$, 
%our model is overly constrained to retain the original sentence while fails to transfer the style.
%\textbf{Third},
%when we replace $L_{cp}$ with $L'_{cp}$ and remove the neural style component simultaneously, 
%the performance of our model further declines.
%This result is in line with our expectation and proves the validity of $L_{cp}$ and the neural style component again.
Table \ref{tab:Ablation} lists the experimental results.
We can observe that  
most variants are significantly inferior to our model in terms of BLEU scores.
Particularly, 
although the \emph{Acc} of some variants increase,
these models may overly change the original content to conduct transfer, 
still resulting in lower BLEU scores.
These results demonstrate the effectiveness of our introduced neural style component, different loss terms and two stage training strategy. As an exception of above observations, when we replace $L_{cp}$ with $L_{cp'}$, the BLEU score increases but the Acc drops significantly. This is due to the fact that $L_{cp'}$ does not discriminate words of different style relevances and overly constrain the model to keep its original content.

\subsection{Case Study}
We conduct case study to understand the advantage of our model. 
Figure \ref{fig:output} displays several instances of input and output sentences with word-level style relevance.
For example, 
according to the word-level style relevance from LRP, 
we can observe that the words of the first input YELP sentence, including ``\emph{slow}'' and ``\emph{lazy}'' are most related to the original style, 
while other words hardly contribute to the style of the whole sentence.
Meanwhile, 
some words of the first output sentence, 
such as ``\emph{really}'', ``\emph{good}'' and ``\emph{friendly}'' are predicted to be most relevant to the target style.
Our model successfully exploits the predicted style relevance and change the word ``\emph{good}'' and ``\emph{friendly}'' while keeping other parts unchanged. In the second informal-to-formal transfer case, besides replacing ``\emph{just}'' with ``\emph{simply}'', our model also appends a ``\emph{.}'' token with high predicted style relevance, which is a sign characterizing formality.

From Figure \ref{fig:output}, we can see that the relevance scores of words in GYAFC are less discriminative compared to those in YELP. Thus, we provide the corpus-level statistics by counting the frequencies of some typical words of GYAFC in Figure \ref{fig:output}, showing that the predicted scores often indicate the distribution of words across different styles. 
``\textit{simply}'' appears 29 and 380 times in the informal and formal sets, respectively. The period token `.' is an indicative marker of text formality since lots of informal sentences ends with no punctuation. 
Besides, ``\textit{your}'', ``\textit{about}'' and ``\textit{this}'' are distributed uniformly across styles. There are 4,590 and 5,357 ``\textit{your}'', 2,489 and 2,353  ``\textit{about}'', 2,062 and 2,211 `\textit{this}''appearing in the informal and formal sets, respectively.

These results are consistent with our intuition, 
verifying the correlation between the predicted style relevance of each word and its actual stylistic outcome.

\section{Conclusion}
This paper has proposed a novel attentional Seq2seq model equipped with a neural style component for unsupervised style transfer.
Using the quantified style relevance from a pre-trained style classifier as supervision information, 
our model is first trained to reconstruct input sentences and repredict the word-level style relevance simultaneously.
Then, equipped with the style component, 
our model can exploit the word-level predicted style relevance for better style transfer.
Experiments on two benchmark datasets prove the superiority of our model over several competitive baselines.

In the future, we plan to adapt variational neural network to refine our style transfer model, which has shown effectiveness in other conditional text generation tasks, such as machine translation \cite{DBLP:conf/emnlp/ZhangXSDZ16,DBLP:conf/aaai/SuWXLHZ18}.

\section*{Acknowledgements}
This work was supported by the Beijing Advanced Innovation Center for Language Resources (No. TYR17002), the National Key R\&D Project of China (No. 2018AAA0101900), the National Natural Science Foundation of China (No. 61672440), and the Scientific Research Project of National Language Committee of China (No. YB135-49).
\bibliography{acl2020}
\bibliographystyle{acl_natbib}
\end{document}